\pdfoutput=1

\documentclass[11pt]{article}

\usepackage{acl}

\usepackage{times}
\usepackage{latexsym}

\usepackage[T1]{fontenc}

\usepackage[utf8]{inputenc}

\usepackage{microtype}

\usepackage{booktabs}
\usepackage{float}
\usepackage{graphicx}
\usepackage{verbatim}
\usepackage{xcolor}
\usepackage{subcaption}
\usepackage{amsmath}
\usepackage{bbm}
\usepackage{soul}

\newcommand{\nascomment}[1]{{\color{blue} [NAS: {#1}]}}

\title{Third-Party Language Model Performance Prediction from Instruction}

\author{Rahul Nadkarni$^{\ast}$ \quad Yizhong Wang$^{*\dagger}$ \quad Noah A.~Smith$^{*\dagger}$ \\
$^\ast$Paul G. Allen School of Computer Science \& Engineering, University of Washington \\
$^\dagger$Allen Institute for Artificial Intelligence \\
  \texttt{\{rahuln,yizhongw,nasmith\}@cs.washington.edu} }

\begin{document}
\maketitle
\begin{abstract}
Language model-based instruction-following systems have lately shown increasing performance on many benchmark tasks, demonstrating the capability of adapting to a broad variety of instructions. However, such systems are often not designed to be transparent about their limitations; a user may easily prompt a model with an instruction without any idea of whether the responses should be expected to be accurate, or if the system is even capable of performing the task. We propose a third party performance prediction framework, where a separate model is trained to predict the metric resulting from evaluating an instruction-following system on a task while assuming access only to its inputs and outputs at inference time. We perform this analysis with a variety of both open and closed instruction-following models as well as multiple performance predictors, and examine the effect of various factors such as model size, number of training tasks, and prompt format. Our findings indicate that third-party performance prediction is very challenging, and much work remains in developing predictors that can automatically reveal the limitations of modern instruction-following natural language processing systems. Our code can be found at \url{https://github.com/rahuln/instr-perf-pred}.

\end{abstract}

\section{Introduction}


Despite much-discussed advances in the capabilities of language model-based systems that follow instructions \citep{Mishra2022, Sanh2022, Wei2022, Ouyang2022, OpenAI2022, OpenAI2023}, the research community lacks an understanding of the limits of these capabilities.  Ideally, purveyors of a technological product would clearly explain to users the limitations of what the system can be used for.\footnote{While documentation such as that of \citet{OpenAI2023} breaks down performance by factors such as task categories, languages, or benchmarks (each benchmark being a collection of many tasks), there is little transparency at any finer granularity (e.g., instruction-level) and no publicly-available tool or mechanism to give users a sense of performance for tasks that do not fit neatly into the documented categories.}  At present, the best a user can do is explore: try out a prompt and see whether the language model can correctly complete the task. We find this state of affairs concerning, because the cost of such tests will fall on the users. Without coordination and information-sharing, different users will make the same explorations and incur unnecessary costs while simultaneously running the risk of relying on systems for tasks which they are incapable of performing adequately.\footnote{A second issue, not addressed here, is that users may not realize that they need to check system output for correctness, and may simply assume that any confident answer from a model can be trusted.  We suspect that this problem will worsen as users explore more and more use cases not anticipated by the builders of the systems and therefore unaddressed by so-called refusal training.}

In this work, we take a step toward empowering users of language-model based systems by proposing a \textbf{third party} approach to predicting model performance at the task level. Consider a user with a particular task in mind. Our proposed task performance predictor takes as input the same prompt the user intends for the language model, and \emph{without querying the language model itself}, offers an estimate of the model's performance on the task.  We instantiate such models by regressing quantitative model performance metrics on natural language task instructions. If successful, such predictors could help users decide among commercial systems, or even opt out of delegating a task to a language model at all.

Our experiments examine how well existing instruction-tuned LMs' performance can be predicted as a function of model size, choice of evaluation metric, amount of training data, and other factors. We find overall that the task is challenging, with the various factors we explore providing little improvement to predictability. Our results underscore how much progress still needs to be made in designing instruction-following natural language systems whose performance can be accurately predicted and made known for the sake of transparency and user safety.

\section{Related Work}
\label{sec:related}

\subsection{Instruction Tuning}

Our work focuses on analyzing the behavior of models that have been trained to follow task instructions. This includes models trained on human-generated instructions and instances \citep{Mishra2022, Sanh2022, Wei2022, Ouyang2022, Wang2022} as well as model-generated data \citep{Wang2023a, Honovich2023, Taori2023, Chiang2023}. We primarily use models trained by \citet{Wang2023b} on a variety of instruction-following datasets and initialized with the publicly available LLaMA family of language models \citep{Touvron2023}. We also explore using the closed models GPT-3.5 \citep{OpenAI2022} and GPT-4 \citep{OpenAI2023}. For more details on our choice of instruction-tuned models, please refer to \S\ref{sec:ims}.

\subsection{Predicting Model Behavior}

Initial work on performance prediction involved training simpler models to predict the performance of larger models as a function of various features, such as model family, model size, task, language, and training procedure \citep{Xia2020, Ye2021, Ye2023}. The primary motivation of that work was to address computational and data constraints -- training separate performance prediction models could alleviate issues where the computational cost of finetuning on all datasets was prohibitive or where scarce data in a particular language or domain prevented finetuning altogether. While much of that work predicted performance at the task level as in our study, its motivations were different and its methods were often implemented by predicting from hand-crafted features related to properties of each dataset (e.g., model parameter count or language features) rather than the text prompt input to the model itself.

More recent efforts to better understand model behavior have analyzed the ability to predict whether or not a model will perform well on a given input, either as determined by the language model itself or a separately-trained model. \citet{Kadavath2022} analyze models to determine whether they can identify examples for which they can generate the correct response, either by prompting for a ``True/False'' label with an instance and one or more generations from the model itself or by finetuning the model as a binary classifier using a dataset of inputs and correctness labels from previous model outputs. \citet{Yao2023} similarly finetune models as binary classifiers to predict whether another model's generated response is correct for tagging, parsing, and semantic parsing tasks. Other work explores models' ability to generate calibrated uncertainty about their responses either through logits, multiple generations, or verbalized expressions of uncertainty, applied to domains such as solving arithmetic problems \citep{Lin2022} or question answering (QA; \citealp{Si2023, Zhou2023, Cheng2023}). Models' ability to verbalize their uncertainty has also been explored for models trained with reinforcement learning with human feedback (RLHF), again using QA datasets \citep{Tian2023}. Notably, all of these efforts examine being able to predict model performance at the \textit{instance level}, whereas we are primarily interested in predicting performance at the \textit{task level} given a task instruction.

The most similar work to ours in spirit is that of \citet{Fu2023}, which also attempts to predict how well a model will perform on a given dataset by training a separate predictor. There are a number of key differences from our efforts: their work focuses on the in-context learning setting using purely pretrained models rather than models specifically trained to follow instructions; they assume access to a set of unlabeled examples for each dataset, rather than just a task instruction; their analysis requires access to model internals by using model output logits to construct a ``confidence profile'' feature vector for each dataset; and they restrict their analysis to a small set of question answering datasets. A modern trend has seen public releases of  models whose internal states are inaccessible, but that can be prompted with a wide variety of user-defined instructions. This situation challenges many of the assumptions made by \citet{Fu2023} and motivates our own work. However, their explorations inspire our efforts, and we attempt to extend their analysis while focusing on task instructions and models trained to follow them.
\section{Methods}

\begin{figure*}[h]
    \centering
    \includegraphics[width=\textwidth]{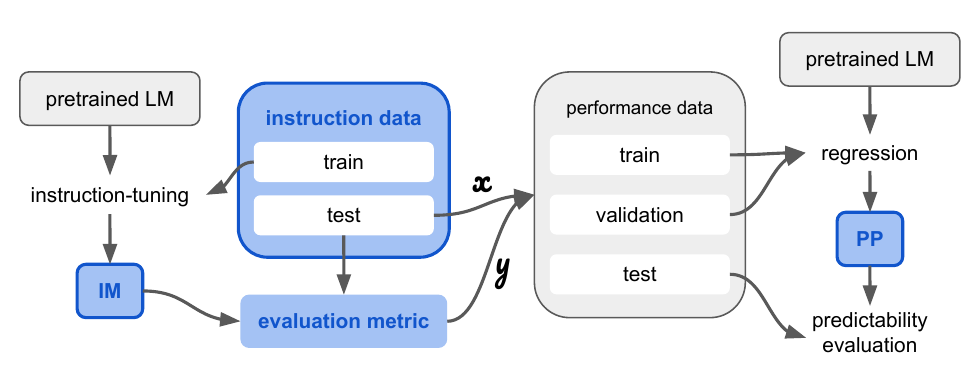}
    \caption{A diagram illustrating our complete analysis pipeline. We begin with a pretrained LM that is instruction-tuned using training tasks from chosen \textbf{instruction data}, resulting in an \textbf{instruction-tuned model (IM)}. The \textbf{IM} is evaluated using the test tasks of the \textbf{instruction data} (not necessarily from the same dataset as the training tasks) and a choice of \textbf{evaluation metric}. Each pair of test task instruction ($\boldsymbol{x}$) and evaluation performance metric value ($y$) is used to construct the performance data, which itself is split into train, validation, and test sets. The train and validation sets are used to train \emph{another} (``third party'') pretrained LM to predict the performance of the \textbf{IM} as a regression model, resulting in the \textbf{performance predictor (PP)}. Finally, the \textbf{PP} is evaluated on the test set of the performance data to determine how well it can predict the performance of the \textbf{IM} on unseen tasks. The sections of the diagram highlighted in \textbf{\color{blue} blue} indicate the components of the pipeline that we vary to determine their effect on performance prediction: the size of the \textbf{IM}, the choice of \textbf{instruction data}, the choice of \textbf{evaluation metric}, and the size and type of \textbf{PP} model.}
    \label{fig:diagram}
\end{figure*}

We begin by describing our complete analysis pipeline, also illustrated as a diagram in Figure~\ref{fig:diagram}. Each of our experiments involves two finetuned language models: one that is trained to follow instructions to perform tasks, which we term the \emph{instruction-tuned model} (IM), and another that is trained to map from an instruction to some measure of the IM's performance on that task, which we call the \emph{performance predictor} (PP).  The IMs considered here are drawn from past work.  To tune a PP, we evaluate a single IM on instructions it was not exposed to during any of its training (including instruction-tuning).  The resulting pairs -- each an instruction $\boldsymbol{x}$ paired with the model's performance score $y$ -- are divided into training, validation, and test sets for the PP.  We train the PP to predict the IM's performance $y$ on instructions $\boldsymbol{x}$ unseen to the IM, and evaluate those predictions on a different set of instructions unseen to either the IM or the PP. We explore a variety of choices for IMs, PPs, and evaluation metrics, which are detailed below.

\subsection{Instruction-Tuned Models (IMs)}
\label{sec:ims}

Our experiments use a range of pretrained LMs that have been finetuned to follow instructions. Part of our goal is to assess the effect of IM size (i.e., parameter count) and choice of instruction-tuning dataset on how well the model's performance can be predicted. To this end, we primarily use LLaMA models \citep{Touvron2023} of various sizes finetuned on a range of instruction-tuning datasets \citep{Wang2023b}. The LLaMA family of models is among the best performing open models, and they have already been used as the basis for a variety of models trained to follow instructions \citep{Taori2023, Chiang2023, Wang2023b}. All models were trained by maximizing the per-token likelihood of the gold output given an instruction and possibly an input, with no additional training procedures such as RLHF \citep{Christiano2017}. While adapting our analysis to models trained using RLHF would be an interesting direction of future work, such models and the data to train them are currently limited relative to models trained with supervised finetuning alone, and our intention is to perform an initial analysis on the least-complicated systems that still demonstrate an ability to follow instructions. For the sake of completeness, we additionally include the closed, API-based models GPT-3.5 (\verb|gpt-3.5-turbo|) and GPT-4, as our goal is to implement a pipeline that is applicable even when access is restricted to model inputs and outputs.

\subsection{Evaluation Metrics}

For each instruction-tuned model, we  perform inference on a separate evaluation set of instructions in order to generate a dataset of the model's behavior ($y$) on unseen instructions ($\boldsymbol{x}$) for training performance prediction models. For each instruction and output pair, we calculate a quantitative evaluation metric that compares the model-generated output to the gold output, in most cases averaging this metric across instances for tasks that have multiple instances. We treat this final metric as the instruction-tuned model's performance for that instruction. We explore two quantitative metrics commonly used when evaluating instruction-tuned models on a broad range of tasks:  ROUGE-L \citep{Lin2004} and Exact Match score. Again, since these automated metrics can be computed by comparing a model-generated output to a gold output, we can apply them to models where we only have access to model generations. Additionally, we briefly compare to using model loss as the metric to predict in \S\ref{sec:predicting-loss}, as an exploration of what we can achieve with performance prediction when additional information  (in this case, the LM's output distribution) is available.

\subsection{Performance Predictors (PP)}

Once we've evaluated an IM on unseen instructions, we use the performance data ($\langle\boldsymbol{x}, y\rangle$ pairs) to build models that predict the IM's performance, and evaluate those predictions on the test subset of the IM performance data. We primarily finetune RoBERTa (both base and large sizes; \citealp{Liu2019}) as the PP, motivated by the goal of having a lightweight separate model that can predict how well an instruction-tuned model will perform without incurring the inference costs of a much larger model. RoBERTa models are trained as regression models by adding a linear layer to the \verb|[CLS]| token at the output layer and training to minimize mean-squared error between the predicted and true evaluation metric for each instruction. In some cases, we also train the base LLaMA model used to build the IM by similarly adding a linear layer to the EOS token at the output layer and adding and updating LoRA adapters \citep{Hu2022} rather than updating all model weights. This is done in an effort to establish an ``upper bound'' of performance prediction that can be achieved while incurring the prohibitive computational cost of using a much larger model. We additionally include a simple baseline of predicting the mean metric value across all training instructions, as a ``lower bound'' to establish whether training a separate predictor model offers any benefit at all.

Unless otherwise specified, PPs are trained and evaluated on the results of evaluating IMs on the Super-NaturalInstructions (SuperNI) test set tasks \cite{Wang2022}. For each experiment, we perform 10 random 80\%/10\%/10\% train/validation/test splits of these tasks and report mean and standard deviation of the performance predictor's root mean squared error (RMSE) in predicting the true evaluation metric from the task instruction. Validation data is used to tune hyperparameters for the PPs, namely batch size and learning rate; for full experiment details, please refer to Appendix~\ref{app:finetuning-details}.
\section{Results}


\subsection{Performance Prediction is Challenging}

\begin{table*}[t]
\centering
\begin{tabular}{llcccccc} \toprule
            \multicolumn{2}{r}{IM:} &                     \textbf{Alpaca-13B} & \textbf{Vicuna-13B} & \textbf{Tülu-13B} & \textbf{GPT-3.5} & \textbf{GPT-4} \\
            \multicolumn{2}{r}{avg.~performance (ROUGE-L):}                     & \textit{45.9} & \textit{44.6} & \textit{61.7} & \textit{53.6} & \textit{63.5}
            \\\midrule
Exact Match & mean          & $25.7_{3.4}$ & $26.4_{3.2}$ & $32.4_{2.3}$ & $30.8_{3.9}$ & $36.4_{3.8}$ \\
            & RoBERTa-base  & $26.2_{4.3}$ & $26.1_{4.9}$ & $33.6_{3.3}$ & $32.5_{5.4}$ & $38.2_{4.1}$ \\
            & RoBERTa-large & $27.4_{5.4}$ & $26.9_{5.4}$ & $34.4_{4.9}$ & $33.0_{5.7}$ & $37.2_{4.5}$ \\
            & LLaMA-13B     & $19.1_{5.3}$ & $18.9_{5.8}$ & $21.4_{3.8}$ & $21.7_{5.6}$ & $22.4_{8.7}$ \\\midrule
ROUGE-L     & mean          & $21.3_{3.5}$ & $20.4_{3.3}$ & $21.4_{1.8}$ & $22.4_{2.9}$ & $24.2_{4.2}$ \\
            & RoBERTa-base  & $22.5_{4.1}$ & $21.2_{3.6}$ & $22.1_{3.8}$ & $23.4_{3.4}$ & $25.6_{4.0}$ \\
            & RoBERTa-large & $22.6_{4.2}$ & $21.2_{3.6}$ & $22.0_{3.7}$ & $23.5_{3.4}$ & $25.7_{3.9}$ \\
            & LLaMA-13B     & $22.2_{4.7}$ & $21.2_{3.8}$ & $21.5_{3.4}$ & $21.6_{3.5}$ & $22.1_{5.1}$ \\
\bottomrule        
\end{tabular}
\caption{Test set RMSE of mean baseline and various PP models finetuned to predict performance from task instruction for various IMs (columns). Subscript shows standard deviation across 10 splits of the SuperNI test tasks.  }
\label{tab:main}
\end{table*}

\begin{figure*}[!t]
    \centering
    \includegraphics[width=\textwidth]{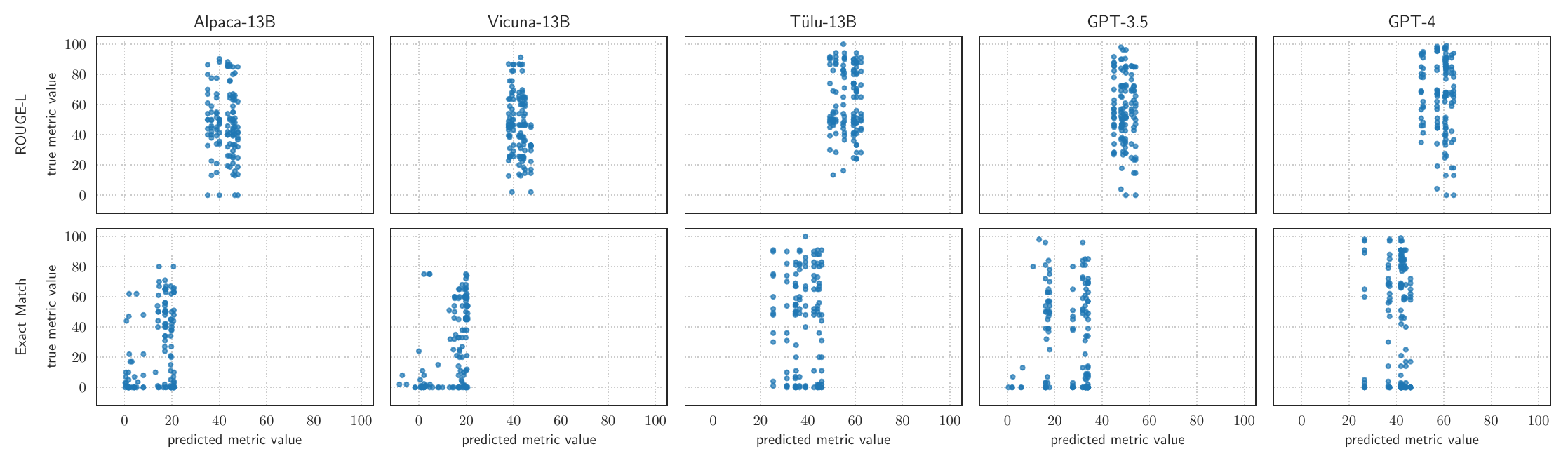}
    \caption{Predicted vs. true metric value when using RoBERTa-large to map from task instruction to performance -- either ROUGE-L (top row) or Exact Match (bottom row) -- for various instruction-following models (columns).}
    \label{fig:scatterplot}
\end{figure*}

Table~\ref{tab:main} shows our main results of predicting performance on SuperNI test set instructions, with mean and standard deviation (subscript) RMSE values across all train-test splits of the SuperNI test tasks. We explore using both Exact Match and ROUGE-L as the target metrics to be predicted for each task, with base and large RoBERTa models as the PP models as well as LLaMA-13B (upper bound) and the simple mean baseline (lower bound). For the IMs, we use the 13B-parameter versions of the Alpaca, Vicuna\footnote{We use the 13B-parameter version of LLaMA finetuned on the ShareGPT dataset by \citet{Wang2023b} as an attempt to replicate the training of Vicuna.}, and Tülu models, as well as GPT-3.5 (\verb|gpt-3.5-turbo|) and GPT-4.

Results demonstrate that performance prediction from task instruction is incredibly difficult, with RMSE values generally around 20 or higher (for metrics in the 0--100 range) across all experimental conditions. Both RoBERTa-base and RoBERTa-large perform comparably to the simple mean baseline, likely indicating that there is little learnable signal within the set of instruction-metric pairs. The lack of a meaningful difference in performance between RoBERTa-base and RoBERTa-large suggests that size of the PP model also makes little difference in performance prediction at the RoBERTa scale. In general, ROUGE-L is more predictable than Exact Match, with consistently lower RMSE values across choices of PPs and IMs. The values below the name of each model indicating performance in ROUGE-L on the SuperNI test tasks show that models which tend to perform better on SuperNI are \emph{less} predictable, exhibiting the worst RMSE values in general.

The results for LLaMA-13B as PP are more promising, with RMSE values that outperform the RoBERTa models as well as the mean baseline when predicting the Exact Match score across all IMs. However, results are still poor, with RMSE values near 20 for both evaluation metrics and all IMs, so even a much larger PP with a more modern base model is unable to predict performance values that are near ground truth. It is additionally worth noting that finetuning LLaMA-13B is a much more prohibitive method of performance prediction, as it involves using a model with around 36$\times$ the number of parameters of RoBERTa-large and incurring comparable inference costs to many of the IMs themselves. As such, we treat the LLaMA-13B performance prediction results as demonstration of a hopeful upper bound while simultaneously underscoring the large room for improvement in developing better-performing and lightweight PP models.

Figure~\ref{fig:scatterplot} shows a more detailed view of a subset of the same results, with scatter plots of the predicted versus true metric value (ROUGE-L or Exact Match) when using RoBERTa-large as the PP for both metrics and all IMs. Results are shown for all train-test splits of the SuperNI test tasks with 12 held-out instructions per random split (10\% of the SuperNI test tasks), resulting in a total of 120 predicted and true performance values for each combination of evaluation metric and IM. This qualitative view of the predictions highlights the fact that RoBERTa-large generally learns to predict roughly the same mean performance value across all tasks within each train-test split of the SuperNI test tasks. These results explain the similarity in performance to the simple mean baseline and demonstrate that the PP model does not learn a meaningful association between instruction and performance.

\subsection{Effect of Various Factors on Predictability}
\label{sec:factors}

We perform more detailed analyses by altering various factors that may affect the behavior of a PP when used to predict the performance of an IM. These factors include the size of the IM, the number of tasks used to train the PP, and the choice of prompt (instruction-only or instruction + 2 positive demonstrations).

\subsubsection{Size of Instruction-tuned Model}

\begin{table}[H]
\centering
\resizebox{\columnwidth}{!}{
\begin{tabular}{lcc} \toprule
                   \multicolumn{1}{r}{PP:}       & \textbf{mean} & \textbf{RoBERTa-large} \\
                          \midrule
LLaMA-7B  (\textit{35.8}) & $21.3_{3.4}$  & $20.7_{4.6}$ \\
LLaMA-13B (\textit{44.6}) & $26.4_{3.2}$  & $27.0_{5.4}$ \\
LLaMA-30B (\textit{44.2}) & $27.5_{2.8}$  & $27.9_{5.3}$ \\
LLaMA-65B (\textit{48.9}) & $30.0_{3.3}$  & $31.7_{5.3}$ \\
\bottomrule        
\end{tabular}
}
\caption{Test set RMSE of predicting Exact Match given task instruction, for various sizes of LLaMA models instruction-tuned on ShareGPT. Model performance in ROUGE-L on the SuperNI test set is given in parentheses next to the name of each model.}
\label{tab:im-size}
\end{table}

We examine the effect of IM size by predicting the performance of various-sized LLaMA models instruction-tuned by \citet{Wang2023b} on the ShareGPT dataset, ranging in scale from 7B to 65B parameters. Results are shown in Table~\ref{tab:im-size}, focusing on the mean baseline and RoBERTa-large PP model as well as the Exact Match metric (results for RoBERTa-base and ROUGE-L are similar). The RMSE values indicate that performance prediction worsens for increasing model size, while model performance on the SuperNI test set improves (ROUGE-L values in parentheses next to each model name). However, the mean baseline also exhibits increasing RMSE values with model scale. This likely suggests that larger, better-performing models are less predictable not because they exhibit more dissimilar behavior on tasks with similar instructions, but because their performance metric values cover a broader range. In any case, we can conclude that scale of the IM alone does not improve performance prediction.

\subsubsection{Number of Training Tasks}

\begin{table}[H]
\centering
\begin{tabular}{lcc} \toprule
       \multicolumn{1}{r}{PP training:}    & \textbf{SuperNI} & \textbf{+ BIG-bench} \\
           \midrule
Alpaca-13B & $27.4_{5.4}$ & $25.4_{6.2}$ \\
Vicuna-13B & $26.9_{5.4}$ & $26.1_{5.5}$ \\
Tülu-13B   & $34.4_{4.9}$ & $34.6_{3.9}$ \\
GPT-3.5    & $33.0_{5.7}$ & $32.3_{6.7}$ \\
GPT-4      & $37.2_{4.5}$ & $37.7_{4.5}$ \\
\bottomrule
\end{tabular}
\caption{Test set RMSE of using RoBERTa-large to predict Exact Match given task instruction, using either just SuperNI instructions or SuperNI and BIG-bench instructions for training the PP model.}
\label{tab:num-tasks}
\end{table}

We examine the effect of increasing the number of tasks / instructions used to train the PP model by including tasks from BIG-bench \citep{Srivastava2023}. Since BIG-bench tasks do not come with instructions, we manually annotate tasks either by converting the task description (if provided) into a SuperNI-style declarative instruction or writing an instruction from scratch when necessary. We additionally filter out a number of tasks, leaving a total of 156 tasks to use as additional training tasks (a full list of included tasks can be found in Appendix~\ref{app:big-bench-tasks}). We perform the same splits of the SuperNI test tasks as before, only adding the BIG-bench tasks to the training split in each case to increase the number of instructions used to train the PP model in an effort to improve its performance.

The results can be found in Table~\ref{tab:num-tasks}. While the inclusion of BIG-bench task instructions represents a nearly 2.5$\times$ increase in the number of training instructions for the PP model, the RMSE values demonstrate that the performance difference is negligible. There are a number of plausible explanations for why this is the case -- in general, BIG-bench tasks are substantially different from SuperNI tasks, including those which deviate from standard language understanding such as decoding encrypted text, performing arithmetic operations, and generating chess moves. The domain shift between the two datasets may explain why the inclusion of the additional instructions did not provide enough meaningful signal for the PP models.

\subsubsection{Prompt Format}

\begin{table}[H]
\centering
\resizebox{\columnwidth}{!}{
\begin{tabular}{lcc} \toprule
        \multicolumn{1}{r}{format:}   & \textbf{instruction} & \textbf{+ 2 demonstrations} \\
           \midrule
Alpaca-13B & $27.4_{5.4}$ & $27.3_{3.7}$ \\
Vicuna-13B & $26.9_{5.4}$ & $29.0_{2.6}$ \\
Tülu-13B   & $34.4_{4.9}$ & $33.6_{2.4}$ \\
GPT-3.5    & $33.0_{5.7}$ & $31.6_{3.3}$ \\
GPT-4      & $37.2_{4.5}$ & $37.8_{3.8}$ \\
\bottomrule
\end{tabular}}
\caption{Test set RMSE of using RoBERTa-large to predict Exact Match given task instruction, using SuperNI tasks with an instruction-only prompt or the instruction with 2 positive demonstrations.}
\label{tab:prompt-format}
\end{table}

We primarily focus on instruction-tuned models evaluated in a zero-shot manner with only an instruction and instance input, as this has become standard practice and reflects how a generic user might interact with an instruction-following system. However, previous work exploring the effect of the prompt format has shown notable improvements in IM performance when including additional information, such as positive demonstrations of the task \citep{Wang2022}. We explore whether this improvement in performance also leads to an improvement in predictability by additionally evaluating all IMs with a prompt format that uses a task instruction with two positive demonstrations, using the positive examples provided with each SuperNI task. The RMSE values resulting from training RoBERTa-large to predict the Exact Match score in this setting, as well as the original instruction-only RMSE values, can be seen in Table~\ref{tab:prompt-format}. While the inclusion of demonstrations improves the performance of the IMs themselves by roughly 2--6 points in Exact Match score, the evaluation results don't lead to the models being more predictable as there are no meaningful differences in values between prompt formats for any model.

\subsection{Predicting Loss}
\label{sec:predicting-loss}

\begin{figure*}[!t]
    \centering
    \includegraphics[width=\textwidth]{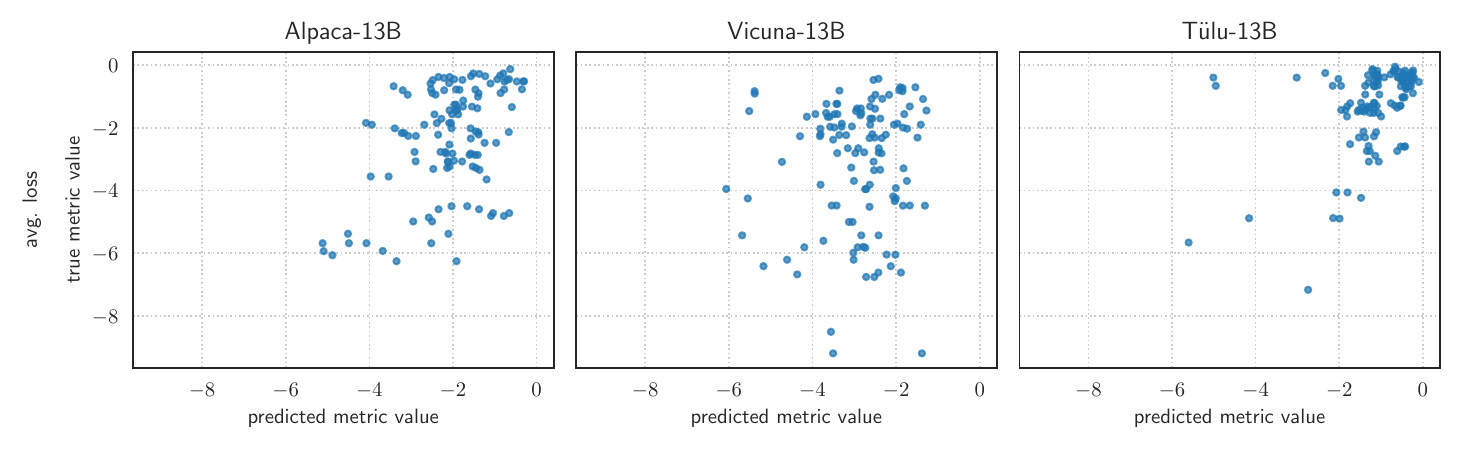}
    \caption{Predicted vs. true loss value when using RoBERTa-large to map from task instruction to loss for various instruction-following models.}
    \label{fig:scatterplot-loss}
\end{figure*}

\begin{table}[H]
\centering
\resizebox{\columnwidth}{!}{
\begin{tabular}{lcc} \toprule
                   \multicolumn{1}{r}{PP:}       & \textbf{mean} & \textbf{RoBERTa-large} \\
                          \midrule
Alpaca-13B  & $1.70_{0.27}$  & $1.51_{0.43}$ \\
Vicuna-13B  & $1.97_{0.46}$  & $2.10_{0.53}$ \\
Tülu-13B    & $1.25_{0.32}$  & $1.10_{0.36}$ \\
\bottomrule        
\end{tabular}
}
\caption{Test set RMSE of predicting cross-entropy loss given task instruction, for various LLaMA-based instruction-following models.}
\label{tab:loss-rmse}
\end{table}

Our core analysis above relied on the use of automated evaluation metrics such as Exact Match and ROUGE-L that can be formulated as functions applied to a pair of generated and gold outputs. This is based on the assumption that we do not have access to model internals or outputs other than generated text, a practical assumption given the increasing use of modern systems that only allow for limited access such as through API requests. However, while such automated metrics can be shown to correlate with other variables such as accuracy on classification tasks \citep{Wang2022}, they come with their own limitations. For instance, for tasks that require more creative or free-form generations from the model, it is possible for the model output and gold label to be semantically equivalent and equally valid responses to the instruction while differing in surface form such that metrics like Exact Match and ROUGE-L are inappropriate \citep{Holtzman2021}. To address this, we perform an additional experiment training PP models to predict the model's \emph{loss} (on the gold output) instead, avoiding token-based comparison between two pieces of text entirely. We do this by evaluating the average per-token cross-entropy loss of the gold label given a prompt (task instruction and input) for each instance in each dataset, averaging across instances to get a single average loss value per task / instruction. We then follow the same regression training procedure to build PP models that predict this value.

Scatter plots similar to Figure~\ref{fig:scatterplot} can be seen in Figure~\ref{fig:scatterplot-loss} (where this time the true and predicted values are this average per-task loss), and the quantitative RMSE values corresponding to these results are in Table~\ref{tab:loss-rmse}. As this analysis requires access to model output distributions to compute the loss, we are limited to the LLaMA-based open models. The plots show the predictions made by RoBERTa-large finetuned to predict average per-task loss, using the same 10 splits of the SuperNI test tasks as in previous results. Qualitatively, these results appear to show greater correlation between the true and predicted values as compared to the metrics considered earlier, indicating that there may be more of a learnable signal for predicting loss. However, the mean RMSE values show that the finetuned RoBERTa-large model still does not outperform the simple mean baseline on average, indicating that performance prediction remains challenging even with access to a quantitative metric not based on token-level comparison between generated and gold outputs. Taken together, our results motivate further work  in identifying quantitative task-level metrics that can be accurately predicted, and perhaps in designing IMs themselves to be more predictable.

\section{Discussion}

Our experiments varied a number of factors that could impact the predictability of an instruction-tuned model's performance on unseen instructions. We broadly summarize our findings below:

\textbf{Performance prediction remains incredibly challenging regardless of setup.} RMSE values remain at 20 or higher for metrics in the range of 0--100, indicating that PP models fail to predict values that are even somewhat close to true performance. This remains true across a variety of open and closed instruction-following models, for multiple automated evaluation metrics. There is also little performance difference between RoBERTa PP models of different scales, and none of them outperform a simple mean baseline. One optimistic result occurs when using LLaMA-13B as the PP model, but performance is still relatively poor and this comes at the cost of scaling up the PP model to the size of the IM.

\textbf{Increasing instruction-tuned model scale, increasing number of training tasks, and adding demonstrations to prompts all fail to improve performance prediction.} The behavior of larger models does not seem to be any more predictable, nor is the behavior of models with access to additional information in the prompt. The typical strategy of increasing the amount of training data available for the finetuned PP model to better learn a prediction signal is also insufficient. The number of instructions remains small, so future work could ascertain whether the problem setup is still limited by the amount of instructions or if there really is no learnable pattern in IM behavior.


\textbf{Predicting cross-entropy loss does not improve performance.} Despite avoiding the issues inherent in using metrics based on token-level comparison between generated and gold outputs, training PP models to predict loss still does not lead to better results than the simple mean baseline.
\section{Conclusion}

Recent NLP systems seem to be able to perform arbitrary tasks given an instruction. Yet we are still not able to understand or explain to users the limitations of these systems such as by reliably predicting their success or failure on new, previously unseen instructions. We take a first step toward this goal by training a separate predictor model to map from a task instruction to the quantified performance of a given instruction-tuned model on that task. Our results show that performance prediction is challenging, with numerous factors like choice of evaluation metric, predictor model size, instruction-following model size, number of training tasks, and prompt format all showing negligible effect on the predictability of instruction-tuned model behavior. Much work remains to be done in designing systems whose limitations can be well-predicted and revealed transparently and freely to their users.
\section{Limitations \& Ethical Considerations}

While we explore the third party performance prediction problem across a variety of factors, there are several constraints that limit our analysis and could be explored in future work. Likely the largest limitation is data -- few datasets exist in the SuperNI style (with multiple tasks each having a declarative instruction and multiple instances for evaluation) that existing models have not already been trained on. Even with the addition of BIG-bench instructions, the resulting dataset is around 250 training instructions, which is still small by most standards and may not provide enough data in general to learn to predict performance. Building more datasets in this format, perhaps by scaling up dataset generation in an automated fashion \citep{Wang2023a, Honovich2023}, could expand this analysis to help overcome the data limitation.

Additionally, using a quantitative, automatic evaluation metric may itself not be appropriate when considering arbitrary tasks, including ones that are creative or based on open-ended generation. Reliable quantitative evaluation in the general instruction-tuned setting is an open challenge. While we attempt to address this with our results predicting cross-entropy loss, the choice of evaluation metric for arbitrary instruction-following tasks remains an open question.

Instruction-tuned model performance can also depend on how the instruction is phrased, and previous work has demonstrated model sensitivity to perturbed or paraphrased instructions \citep{Zhao2021, Webson2022}. Driven by assumptions that most users will not ``engineer'' instructions extensively, our experiments only consider a single instruction per task, and redefining performance based on multiple instructions per task (i.e., more ``task-specific'' rather than ``instruction-specific'' behavior) may lead to other interesting results.
\section*{Acknowledgements}

We would like to thank the members of Noah's ARK research group for their helpful feedback on this work. This research was supported in part by the Office of Naval Research under MURI grant N00014-18-1-2670.

\bibliography{anthology,custom}
\bibliographystyle{acl_natbib}

\appendix
\section*{Appendix}

\section{Finetuning Details}
\label{app:finetuning-details}

We use the Huggingface \verb|transformers| library to run all experiments \citep{Wolf2020}. Details for finetuning RoBERTa-base and RoBERTa-large performance predictors can be found in Table~\ref{tab:settings-roberta}, while details for the LLaMA-13B performance predictors can be found in Table~\ref{tab:settings-llama}. We perform a hyperparameter search over batch size and learning rate for all finetuning experiments. RoBERTa models are finetuned by updating all parameters, while LLaMA models are finetuned by adding and updating LoRA adapters \cite{Hu2022}. All performance predictors use an additional linear layer applied to either the \verb|[CLS]| token (for RoBERTa) or the EOS token (for LLaMA) at the last layer to make the final prediction. All models were evaluated on the validation set of instructions at every epoch, with early stopping performed based on validation set RMSE.

\begin{table}[H]
\centering
\resizebox{\columnwidth}{!}{
\begin{tabular}{lc} \toprule
\textbf{Hyperparameter} & \textbf{Assignment} \\
\midrule
number of epochs        & 20 \\
batch size              & \{4, 8, 16\} \\
maximum learning rate   & \{1e-5, 5e-5, 1e-4, 5e-4\} \\
optimizer               & AdamW \\
epsilon                 & 1e-8 \\
betas                   & (0.9, 0.999) \\
learning rate schedule  & constant \\
weight decay            & 0 \\
warmup proportion       & none \\
learning rate decay     & none \\
\bottomrule
\end{tabular}}
\caption{Experiment settings for finetuning RoBERTa performance predictor models.}
\label{tab:settings-roberta}
\end{table}

\begin{table}[H]
\centering
\resizebox{\columnwidth}{!}{
\begin{tabular}{lc} \toprule
\textbf{Hyperparameter} & \textbf{Assignment} \\
\midrule
number of epochs        & 20 \\
batch size              & \{8, 16, 32\} \\
maximum learning rate   & \{1e-5, 2e-5, 5e-5, 1e-4\} \\
optimizer               & AdamW \\
epsilon                 & 1e-8 \\
betas                   & (0.9, 0.999) \\
learning rate schedule  & linear warmup \\
weight decay            & 0 \\
warmup proportion       & 0.03 \\
learning rate decay     & linear \\
LoRA rank               & 256 \\
LoRA alpha              & 256 \\
LoRA dropout            & 0.05 \\
\bottomrule
\end{tabular}}
\caption{Experiment settings for finetuning LLaMA performance predictor models.}
\label{tab:settings-llama}
\end{table}

\section{Details on BIG-bench}
\label{app:big-bench-tasks}

The full list of tasks included from BIG-bench can be found in Table~\ref{tab:big-bench-tasks}.

\begin{table*}[htbp]
\centering
\begin{tabular}{@{}p{\textwidth}@{}}
\toprule
\textbf{Tasks} \\ \midrule
abstract\_narrative\_understanding, anachronisms, analogical\_similarity, analytic\_entailment, arithmetic, ascii\_word\_recognition, authorship\_verification, auto\_categorization, auto\_debugging, bbq\_lite\_json, bridging\_anaphora\_resolution\_barqa, causal\_judgment, cause\_and\_effect, checkmate\_in\_one, chess\_state\_tracking, chinese\_remainder\_theorem, cifar10\_classification, code\_line\_description, codenames, color, common\_morpheme, conceptual\_combinations, conlang\_translation, contextual\_parametric\_knowledge\_conflicts, crash\_blossom, crass\_ai, cryobiology\_spanish, cryptonite, cs\_algorithms, dark\_humor\_detection, date\_understanding, disambiguation\_qa, discourse\_marker\_prediction, disfl\_qa, dyck\_languages, elementary\_math\_qa, emoji\_movie, emojis\_emotion\_prediction, empirical\_judgments, english\_proverbs, english\_russian\_proverbs, entailed\_polarity, entailed\_polarity\_hindi, epistemic\_reasoning, evaluating\_information\_essentiality, fact\_checker, fantasy\_reasoning, few\_shot\_nlg, figure\_of\_speech\_detection, formal\_fallacies\_syllogisms\_negation, gem, gender\_inclusive\_sentences\_german, general\_knowledge, geometric\_shapes, goal\_step\_wikihow, gre\_reading\_comprehension, hhh\_alignment, hindi\_question\_answering, hindu\_knowledge, hinglish\_toxicity, human\_organs\_senses, hyperbaton, identify\_math\_theorems, identify\_odd\_metaphor, implicatures, implicit\_relations, intent\_recognition, international\_phonetic\_alphabet\_nli, international\_phonetic\_alphabet\_transliterate, intersect\_geometry, irony\_identification, kanji\_ascii, kannada, key\_value\_maps, known\_unknowns, language\_games, language\_identification, linguistics\_puzzles, logic\_grid\_puzzle, logical\_args, logical\_fallacy\_detection, logical\_sequence, mathematical\_induction, matrixshapes, metaphor\_boolean, metaphor\_understanding, minute\_mysteries\_qa, misconceptions, mnist\_ascii, modified\_arithmetic, moral\_permissibility, movie\_dialog\_same\_or\_different, movie\_recommendation, mult\_data\_wrangling, navigate, nonsense\_words\_grammar, novel\_concepts, object\_counting, odd\_one\_out, operators, paragraph\_segmentation, parsinlu\_qa, parsinlu\_reading\_comprehension, penguins\_in\_a\_table, periodic\_elements, persian\_idioms, phrase\_relatedness, physical\_intuition, physics, physics\_questions, play\_dialog\_same\_or\_different, polish\_sequence\_labeling, presuppositions\_as\_nli, qa\_wikidata, question\_selection, real\_or\_fake\_text, reasoning\_about\_colored\_objects, repeat\_copy\_logic, rephrase, riddle\_sense, ruin\_names, salient\_translation\_error\_detection, scientific\_press\_release, semantic\_parsing\_in\_context\_sparc, semantic\_parsing\_spider, sentence\_ambiguity, similarities\_abstraction, simp\_turing\_concept, simple\_ethical\_questions, simple\_text\_editing, snarks, social\_iqa, social\_support, sports\_understanding, strange\_stories, strategyqa, sufficient\_information, suicide\_risk, swahili\_english\_proverbs, swedish\_to\_german\_proverbs, symbol\_interpretation, temporal\_sequences, tense, timedial, topical\_chat, tracking\_shuffled\_objects, understanding\_fables, undo\_permutation, unit\_conversion, unit\_interpretation, vitaminc\_fact\_verification, what\_is\_the\_tao, which\_wiki\_edit, winowhy, word\_sorting, word\_unscrambling \\
\bottomrule
\end{tabular}
\caption{List of BIG-bench tasks included when training performance predictors on additional tasks.}
\label{tab:big-bench-tasks}
\end{table*}

\end{document}